\documentclass{article}

\PassOptionsToPackage{numbers, compress}{natbib}
\usepackage[preprint]{neurips_2022}

\usepackage[utf8]{inputenc} % allow utf-8 input
\usepackage[T1]{fontenc}    % use 8-bit T1 fonts
\usepackage{hyperref}       % hyperlinks
\usepackage{url}            % simple URL typesetting
\usepackage{booktabs}       % professional-quality tables
\usepackage{amsfonts}       % blackboard math symbols
\usepackage{nicefrac}       % compact symbols for 1/2, etc.
\usepackage{microtype}      % microtypography
\usepackage{xcolor}         % colors
\usepackage{graphicx}
\usepackage{wrapfig}
\usepackage[table]{colortbl}
\usepackage[compact]{titlesec}
\definecolor{green}{RGB}{0,195,0}

\title{Exploring Visual Prompts \\for Adapting Large-Scale Models}
\author{%
  Hyojin Bahng\\
  MIT CSAIL\\
  \texttt{bahng@mit.edu} \\
   \And
   Ali Jahanian\thanks{Equal contribution.}  \\
   MIT CSAIL\\
   \texttt{jahanian@mit.edu} \\
   \AND
   Swami Sankaranarayanan\footnotemark[1]\\
   MIT CSAIL\\
   \texttt{swamiviv@mit.edu} \\
   \And
   Phillip Isola\\
   MIT CSAIL\\
   \texttt{phillipi@mit.edu} \\
}

\begin{document}

\maketitle

\begin{abstract}
We investigate the efficacy of \emph{visual prompting} to adapt large-scale models in vision. Following the recent approach from prompt tuning and adversarial reprogramming, we learn a single image perturbation such that a frozen model prompted with this perturbation performs a new task. Through comprehensive experiments, we demonstrate that visual prompting is particularly effective for CLIP and robust to distribution shift, achieving performance competitive with standard linear probes. We further analyze properties of the downstream dataset, prompt design, and output transformation in regard to adaptation performance. The surprising effectiveness of visual prompting provides a new perspective on adapting pre-trained models in vision. Code is available at~\url{https://hjbahng.github.io/visual_prompting/}.
\end{abstract}

%\vspace{1pt}

\section{Introduction}

When we humans learn a new task, we tend to start from our current knowledge base and extrapolate thereof. A child who is starting to speak and comprehend sentences quickly develops the ability to parse the emotional context that accompanies a sentence. For example, the sentence “I missed the school bus” carries a particular emotion such that if followed by “I felt so \texttt{[MASK]}”, the child can provide an appropriate emotion word. This paradigm, aptly named \emph{prompting}, has recently been popularized in NLP, where large pre-trained language models are adapted to new tasks by converting the downstream dataset into the format of the pre-training task. Without updating any of its parameters, the language model uses its existing knowledge base to fill in the mask in the provided prompt, hence becoming an expert in the new task. Currently, prompting methods are dominantly NLP-specific~\cite{brown2020language, schick2020exploiting, schick2020s, shin2020autoprompt, gao2020making, liu2021gpt, li2021prefix, qin2021learning, hambardzumyan2021warp, zhong2021factual, han2021ptr, lester2021power}, despite the fact that the framework serves a general purpose: adapt a frozen pre-trained model by modifying the \emph{data space}. Considering the generality, can we create prompts in the form of \emph{pixels}? Broadly, can we steer frozen visual models to solve a new task by modifying pixel space? 

%Our primary research question is: does prompting work in \emph{pixel space}? In particular, how far can we improve adaptation to a new task \emph{solely by modifying pixels}? 

Adversarial reprogramming~\cite{elsayed2018adversarial} is a class of adversarial attacks where input perturbations repurpose a model to perform a task chosen by the adversary. Despite having different terms\footnote{For convenience, we unify the term and use ``visual prompt'' to denote any pixel-space modification to the input image for model adaptation.} and motivations, this input perturbation essentially acts as a \emph{visual prompt} --- it adapts a model to new tasks by modifying pixels. Existing methods~\cite{elsayed2018adversarial, kloberdanz2021improved, chen2021adversarial, randazzo2021adversarial}, however, have focused on adversarial goals or demonstrated limited application to relatively small-scale datasets and models. Having originated from different communities, adversarial reprogramming and prompting share the general idea~\cite{chen2022model,liu2021pre}: perform \emph{data-space adaptation} by transforming the input (i.e., prompt engineering) and/or output (i.e., answer engineering). 
% Inspired by the success of natural language prompting, we revisit visual prompts as a practical adaptation method for modern vision models.

\begin{figure}[t]
% \vskip 0.1in
\centering
\includegraphics[width=1\linewidth]{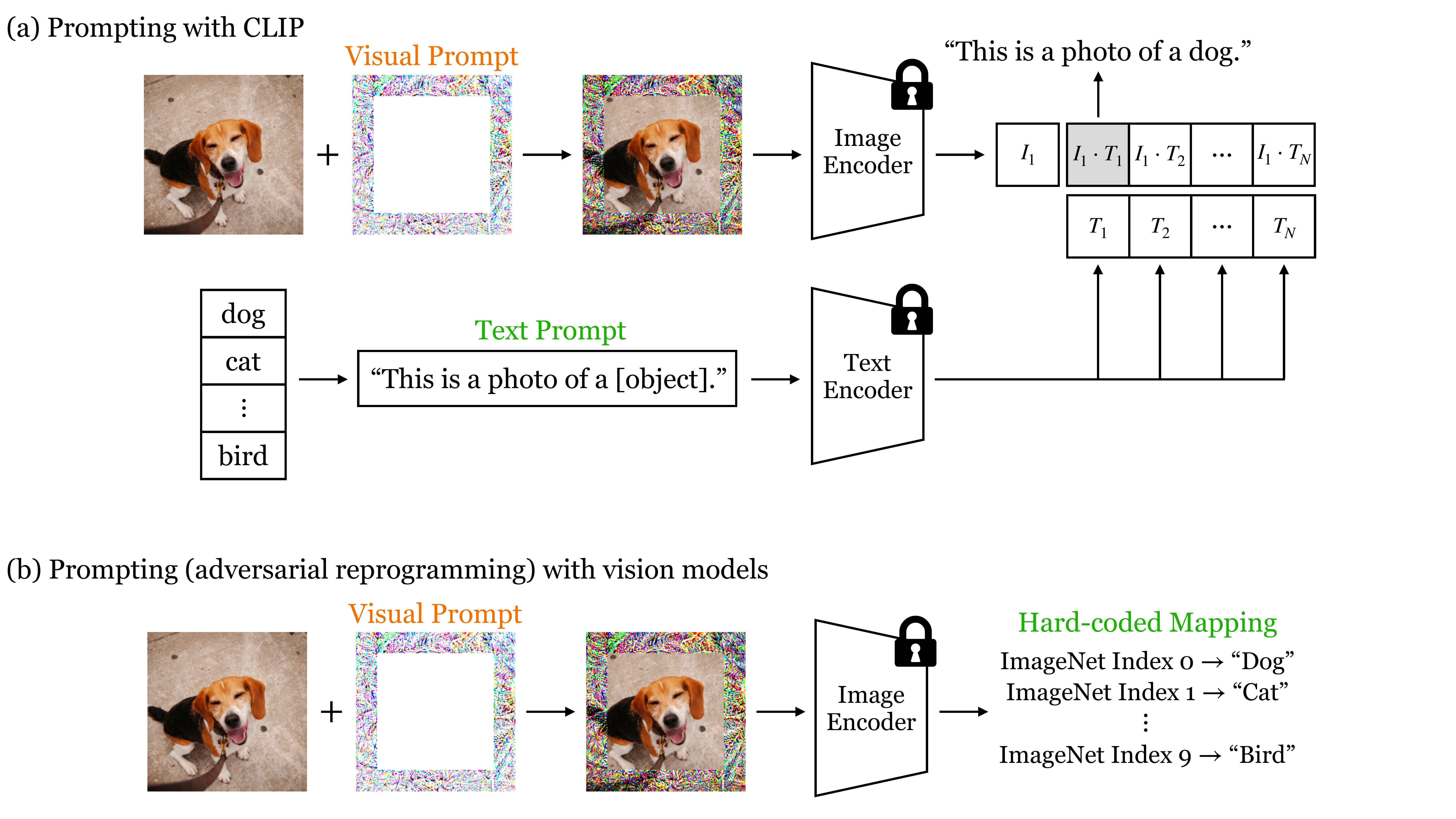}
\caption{\textbf{Prompting for CLIP and vision models.} Prompting transforms the \textcolor{orange}{input} and/or \textcolor{green}{output} of the downstream dataset into the format of the pre-trained task. We learn a single \textbf{visual prompt} via backpropagation to transform all input images. We map the model outputs to downstream labels by using a (a) discrete text prompt for CLIP and (b) hard-coded mapping for vision models.
} 
\label{clip_vs_vision}
\vskip -0.1in
\end{figure}

% In this work, we aim to investigate the efficacy of \emph{visual prompts} for adapting large-scale models in vision.
Inspired by the success of natural language prompting, we aim to investigate the efficacy of \emph{visual prompting} for adapting large-scale models in vision. As pixel space is inherently continuous, we follow the recent approach that treats prompts as a continuous task-specific vector~\cite{elsayed2018adversarial,li2021prefix,hambardzumyan2021warp,lester2021power}. We learn a single image perturbation (i.e., ``soft prompt'') via backpropagation while having the model parameters frozen. We map the model outputs to downstream labels by using a discrete text prompt for CLIP~\cite{radford2021learning} and hard-coded mapping for vision models (Figure~\ref{clip_vs_vision}). 
% This allows adaptation to new tasks without \emph{any} modification to the model which is in stark contrast with current de facto adaptation methods in computer vision...

How is visual prompting different from existing adaptation methods? Currently in vision, standard adaptation methods are fine-tuning and linear probe. Both approaches require some level of access to the model: entire parameters in the case of fine-tuning and model outputs (usually activations at the penultimate layer) in the case of linear probe. In contrast, visual prompting adapts the input to a model. After acquiring the visual prompt, it does not require model access at test time. This opens up
unique applications~\cite{salman2021unadversarial}; input-space adaptation puts control in the hands of the end-user of the system. For instance, a pedestrian could wear a visual prompt that improves their visibility to cars, without having access to the car itself, nor its vision system. 

We conduct comprehensive experiments across four pre-trained models and 15 image classification datasets. We demonstrate that visual prompting is surprisingly effective for CLIP~\cite{radford2021learning} and robust to distribution shift, achieving performance competitive with, and sometimes beyond, standard linear probes. We further analyze what properties of the downstream dataset, prompt design, and output transformation affect performance. Note that our goal is not to achieve the state-of-the-art performance on specific tasks, but instead to broadly explore a new paradigm for visual adaptation. The surprising effectiveness of visual prompting provides a new perspective on how to adapt and use pre-trained models in vision.

% \vspace{-15pt}

\section{Related Work}
\vspace{-2pt}
\label{prob_def}
\subsection{Natural Language Prompting}
Our investigation is inspired by the recent success in natural language prompting. Prompting in NLP reformulates the downstream dataset into a (masked) language modeling problem, so that a frozen language model directly adapts to a new task without updating any parameters. A prompt consists of constructing a task-specific \emph{template} (e.g., ``I felt so \texttt{[MASK]}'') and \emph{label words} (e.g., ``happy/horrible'') to fill in the blank~\cite{gao2020making}. However, hand-crafting the right prompt requires domain expertise and a significant amount of effort.

Prefix tuning~\cite{li2021prefix} or prompt tuning~\cite{lester2021power} mitigates this problem by learning a ``soft prompt'' via backpropagation, while having the model parameters fixed. Prefix tuning learns a task-specific continuous vector (i.e., prefix) that allows language models to adapt to various generation tasks. While prefix tuning prepends the prefix to each encoder layer, prompt tuning further simplifies by only prepending tunable tokens to the input. When applied to large models with billions of parameters, a properly optimized prompt achieves competitive performance to fine-tuning the entire model, while significantly reducing memory usage and per-task storage. As prompts in pixel space are inherently continuous, we follow this line of work and optimize the pixels directly. 

\subsection{Prompting with Images} 
There have been initial approaches that attempt to prompt with images. Similar to prefix tuning, Frozen~\cite{tsimpoukelli2021multimodal} creates a image-conditional prompt by training a vision encoder using gradients from a frozen language model. The images are represented as a continuous embedding from the vision encoder and used as a \emph{visual prefix} to allow frozen language models to perform multi-modal tasks. CPT~\cite{yao2021cpt} converts visual grounding into a fill-in-the-blank problem by creating visual prompts with colored blocks and color-based textual prompts. However, both of these approaches focus on extending the capabilities of a \emph{language}-based model. On the other hand, we focus on investigating the efficacy of prompting for visual representations and image classification datasets. In other words, we assume that the pre-trained model consists of a visual encoder and focus on reformulating image datasets. Visual prompt tuning~\cite{jia2022visual} is concurrent work that proposes visual prompts specific to Vision Transformers~\cite{dosovitskiy2020image}. It uses deep prompt tuning~\cite{li2021prefix, qin2021learning, liu2021p} by prepending a set of tunable parameters to each Transformer encoder layer.

\subsection{Adversarial Reprogramming and Unadversarial Examples}ß
Adversarial reprogramming~\cite{elsayed2018adversarial} is a type of adversarial attack where a single, class-agnostic perturbation reprograms a model to perform a new task chosen by the attacker. Despite its adversarial goal, the framework essentially serves the same purpose as prompting: adapt a frozen model to new tasks by modifying the input and/or output of the downstream dataset. However, existing methods in vision~\cite{chen2021adversarial,kloberdanz2021improved,randazzo2021adversarial} are designed to achieve an adversarial goal or demonstrate limited application to small-scale vision models and simple datasets. 
Similarly, unadversarial examples~\cite{salman2021unadversarial} aim to increase performance on the (pre-)trained task. It learns an image perturbation that improves performance on a specific class (i.e., class-conditional). In our work, we revisit adversarial reprogramming as a form of visual prompt and investigate its efficacy in adapting large-scale models in vision.

\subsection{Adapting Pre-trained Models in Vision} 
\begin{figure}[t]
% \vskip 0.2in
\centering
\includegraphics[width=\textwidth]{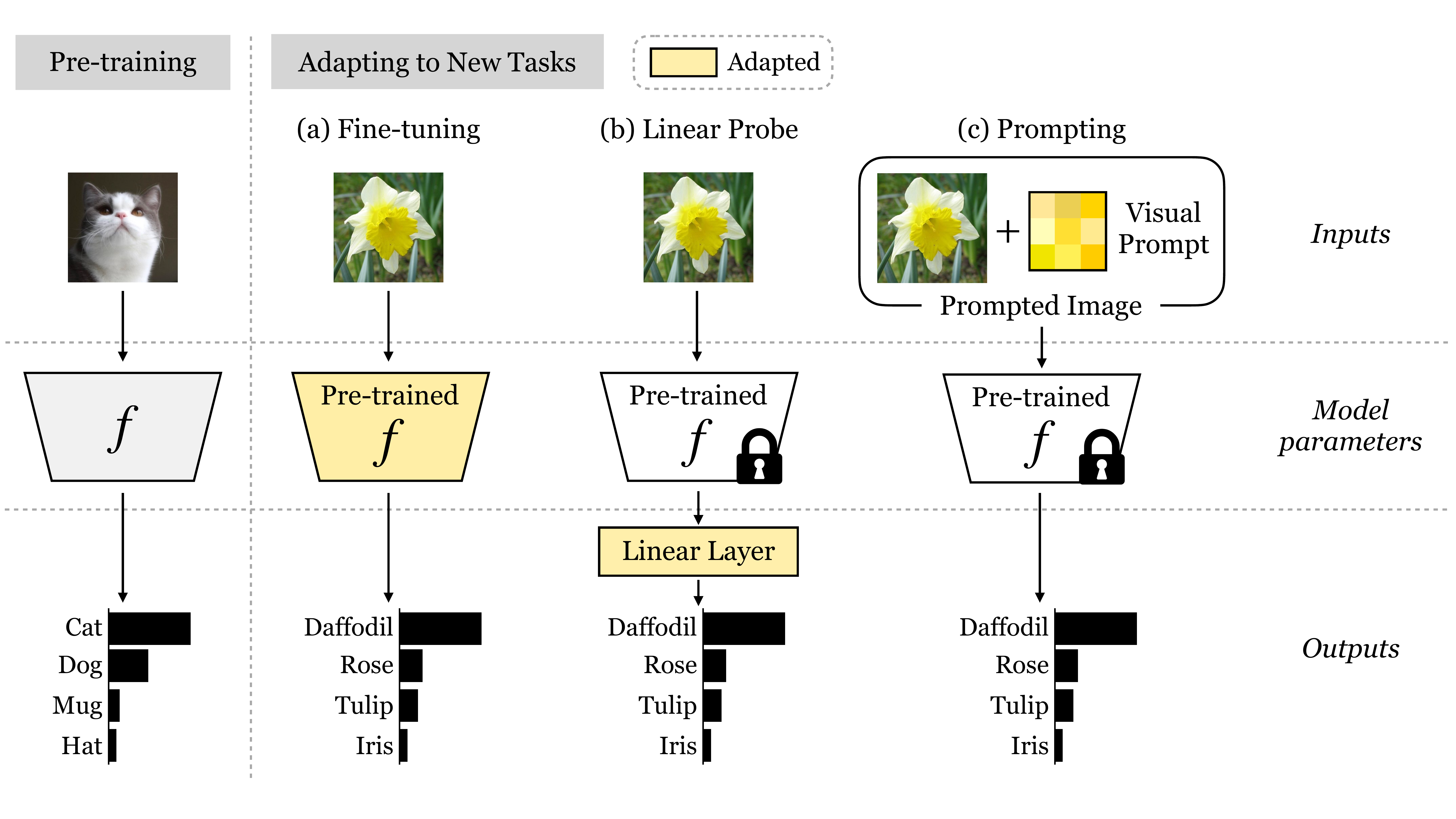}
\caption{\textbf{Methods for adapting pre-trained models to downstream tasks.} (a) Fine-tuning adapts the entire model parameters. (b) Linear probes adapt the model outputs (usually activations at the penultimate layer) by learning a linear layer. (c) Prompting adapts the (downstream) dataset by reformulating the input and/or output.}
\label{adapt_overview}
% \vskip -0.2in
\end{figure}

Figure~\ref{adapt_overview} provides a summary of different methods for adapting a pre-trained model. Fine-tuning and linear probing are highly flexible in their usage: they can be used to adapt the model to a new domain of inputs or to a new task with different output semantics. However, they also require some level of access to the model: parameters in the case of fine-tuning and model outputs (usually activations at the penultimate layer) in the case of linear probes. Domain adaptation is an interesting alternative to model adaptation in that it only modifies the \textit{inputs} to the model using techniques such as image-to-image translation~\cite{zhu2017unpaired,hoffman2018cycada}. Like domain adaptation, visual prompting also modifies the inputs to a model. Therefore, once the end user has found the visual prompt, it does not require having control over the model itself at test time. This opens up unique applications; for example, users can feed domain-adapted images to online APIs that can only be manipulated via their inputs. Domain adaptation focuses on adapting a source domain to \textit{look like} a target domain, requiring both source and target datasets available at hand. On the other hand, we demonstﬁrate that visual prompting can steer model in more arbitrary ways; for example, a model that performs one classification task can be adapted to perform \textit{an entirely different classification task}, with new output semantics, just by perturbing the input pixels. Also, whereas domain adaptation methods are typically input-conditional, the visual prompts we explore in this paper are \textit{fixed} (i.e., input-agnostic) across an entire dataset, as in NLP where the same natural language prompt is added to all model queries.

\section{Methods}
\label{method}
Under different terms, prompting and adversarial reprogramming serve the same purpose: data-space adaptation. They generally consist of two stages~\cite{chen2022model,liu2021pre}: input transformation and output transformation. The goal of the input transformation (or prompt engineering) is to design a proper prompt that specifies the task which is applied to the input. The goal of the output transformation (or answer engineering) is to map the model's output/answer to the target label. We introduce different design choices for vision and vision-language models according to their pre-trained task.

\subsection{Pre-trained Models}
Prompts in pixel form can essentially be applied to any visual representation. Therefore, we select three vision models and one vision-language model: Instagram-pretrained ResNeXt (Instagram)~\cite{mahajan2018exploring}, Big Transfer (BiT-M)~\cite{kolesnikov2020big}, ResNet trained on ImageNet-1k (RN50)~\cite{he2016deep,deng2009imagenet}, and CLIP~\cite{radford2021learning}. 
Vision models are trained to predict a fixed set of predetermined classes and typically require learning a separate layer to predict unseen classes. In contrast, CLIP is a vision-language model that is able to perform flexible zero-shot transfer to unseen classes using text prompts. 
We summarize the pre-trained model details in the Appendix. We select models across varying input modalities, pre-trained dataset size, and model architecture to evaluate the practical utility of visual prompts. For Instagram-pretrained ResNeXt, we use the model additionally fine-tuned on ImageNet-1k. 

\subsection{Input Transformation}
There can be several ways of designing a visual prompt. As pixel space is less discrete compared to natural language, it is difficult to handcraft prompts as in NLP (e.g., ``a photo of a \texttt{[LABEL]}'' for image classification). In fact, it is unclear what type of visual context is useful for each downstream task (e.g., what visual information would be useful for specifying satellite image classification?). Intuitively, a visual prompt does not necessarily need to be interpretable to humans; it’s a visual cue that aids the decision of a machine learning model. Thus, let the model optimize the visual context! We follow a simple gradient-based approach~\cite{elsayed2018adversarial,hambardzumyan2021warp,li2021prefix,lester2021power} where we directly optimize the visual prompt via backpropagation. 

\subsubsection{Prompt Tuning}
Given a frozen pre-trained model $F$ and a downstream task dataset $\mathcal{D}=\{(x_1,y_1), \dots, (x_m,y_m)\}$, our objective is to learn a single, task-specific visual prompt $v_\phi$ parameterized by $\phi$. The prompt is added to the input image to form a prompted image $x+v_\phi$. During training, the model maximizes the likelihood of the correct label $y$, 
\begin{equation}
\max_\phi P_{\theta;\phi}(y|x+v_\phi), 
\end{equation}
while the gradient updates are applied only to the prompt parameters $\phi$ and the model parameters $\theta$ remain frozen. During evaluation, the optimized prompt is added to all test-time images, 
\begin{equation}
X_\mathrm{test}=\{x_1+v_\phi, \dots, x_n+v_\phi\},
\end{equation}
which are then processed through the frozen model $F$. 

Note that our goal is to explore visual prompts as a practical adaptation method. Therefore, we do not necessitate any adversarial constraint of making the perturbations imperceptible. Also, adversarial reprogramming assumes the downstream dataset to be lower-resolution than the pre-trained dataset, such that the input perturbation is padded around the downstream dataset. In real-world applications, the downstream dataset can have varying resolutions. Thus, we resize every dataset to the input size of the pre-trained model and add the prompt directly to the input region.

\subsubsection{Prompt Design}
\label{prompt_design}
There are several ways to design a visual prompt in terms of template and size. We explore three visual templates: pixel patch at random location, pixel patch at fixed location, and padding. We explore various prompt sizes $p$, where the actual number of parameters is $Cp^2$ for patches and $2Cp(H+W-2p)$ for padding, where $C$, $H$, $W$ are the image channels, height and width respectively. Section~\ref{prompt_ablation} shows that padding with $p=30$ achieves the best performance over other design choices. We use this as default for all our experiments.  

\subsection{Output Transformation}
To map model outputs to the target label, we take a different approach for vision models and CLIP. Standard vision models treat image classes as a numeric id (e.g., ``cat'' is mapped to ``index 1''). We use a hard-coded mapping~\cite{elsayed2018adversarial} and arbitrarily map downstream class indices to pre-trained class indices, discarding unassigned indices for loss computation. For CLIP, a vision-language model, we utilize text prompts~\cite{radford2021learning} as our output transformation function. Image classes are represented by text (e.g., ``cat'') which are then prompted (e.g., ``a photo of a \texttt{[object]}'') to specify context of the downstream task. Note that we use a single, fixed text prompt (see Appendix) and only optimize the visual prompt. We follow the protocol for CLIP zero-shot transfer and calculate cosine similarity of the embeddings for every class, which is normalized into a probability distribution via softmax. The class with the highest probability is chosen as the model output. The full overview for vision models and CLIP is illustrated in Figure~\ref{clip_vs_vision}.

\subsection{Implementation Details}
To learn the visual prompt, the objective function for CLIP is identical to its evaluation setting, i.e., we only compute cross entropy loss over images, where a set of prompted text strings is processed through the text encoder to produce weights of a linear classifier~\cite{radford2021learning}. For vision models, we compute cross entropy loss over new class indices. For all experiments, we use the padding template with prompt size of 30. All images are resized to $224\times224$ to match the input size of pre-trained models, and preprocessed identical to the evaluation setting of each model. We find that closely following the pre-trained model's evaluation setting is important for learning a good prompt. All visual prompts are trained for 1,000 epochs. We use SGD with a learning rate of 40, which is decayed using cosine schedule~\cite{loshchilov2016sgdr}. We use a batch size of 256 for CLIP, 128 for BiT-M and RN50, and 32 for Instagram.

\section{Experimental Setup} 
\label{exp}
\subsection{Datasets}
To evaluate how well visual prompts adapt a model to new tasks, we measure performance across 12 datasets: CIFAR100, CIFAR10~\cite{krizhevsky2009learning}, Flowers102~\cite{nilsback2008automated}, Food101~\cite{bossard2014food}, EuroSAT~\cite{helber2019eurosat}, SUN397~\cite{Xiao:2010}, DTD~\cite{cimpoi2014describing}, UCF101~\cite{soomro2012ucf101}, SVHN~\cite{netzer2011reading}, OxfordPets~\cite{parkhi2012cats}, Resisc45~\cite{cheng2017remote}, and CLEVR~\cite{johnson2017clevr}. 
We also measure robustness to distribution shift, i.e., training distribution differs from the test distribution, by evaluating on three image classification datasets in WILDS~\cite{wilds2021}: Camelyon17~\cite{bandi2018detection}, FMoW~\cite{christie2018functional}, and iWildCAM~\cite{beery2020iwildcam}. For Camelyon17, training and test sets comprise tissue patches from different hospitals. For FMoW, training and test sets are from different regions and years. Finally, iWildCAM consists of photos from disjoint sets of camera traps. Note that we learn the visual prompt on the training set and evaluate its performance on the test set.

\subsection{Baseline Methods}
To measure how visual prompting performs compared to existing adaptation methods (Figure~\ref{adapt_overview}), we compare fine-tuning, linear probes, and text prompting (i.e., zero-shot transfer). Fine-tuning and linear probe are standard adaptation methods in vision. Fine-tuning update the entire model parameters during adaptation. Linear probe is a lightweight alternative which adapts the model outputs (usually activations at the penultimate layer) by learning a linear layer, while having the model parameters frozen. For text prompting, we use ``This is a photo of a \texttt{[LABEL]}'' as default. For CLEVR, we use ``This is a photo of \texttt{[LABEL]} objects'', with class label ``three'' to ``ten''. For Camelyon17, we use ``a tissue region \texttt{[LABEL]} tumor'', with class label ``containing'' and ``not containing''.

\begin{figure}[t]
    \begin{minipage}[t]{.49\linewidth}
    \centering
            \centering
    \includegraphics[width=1\linewidth]{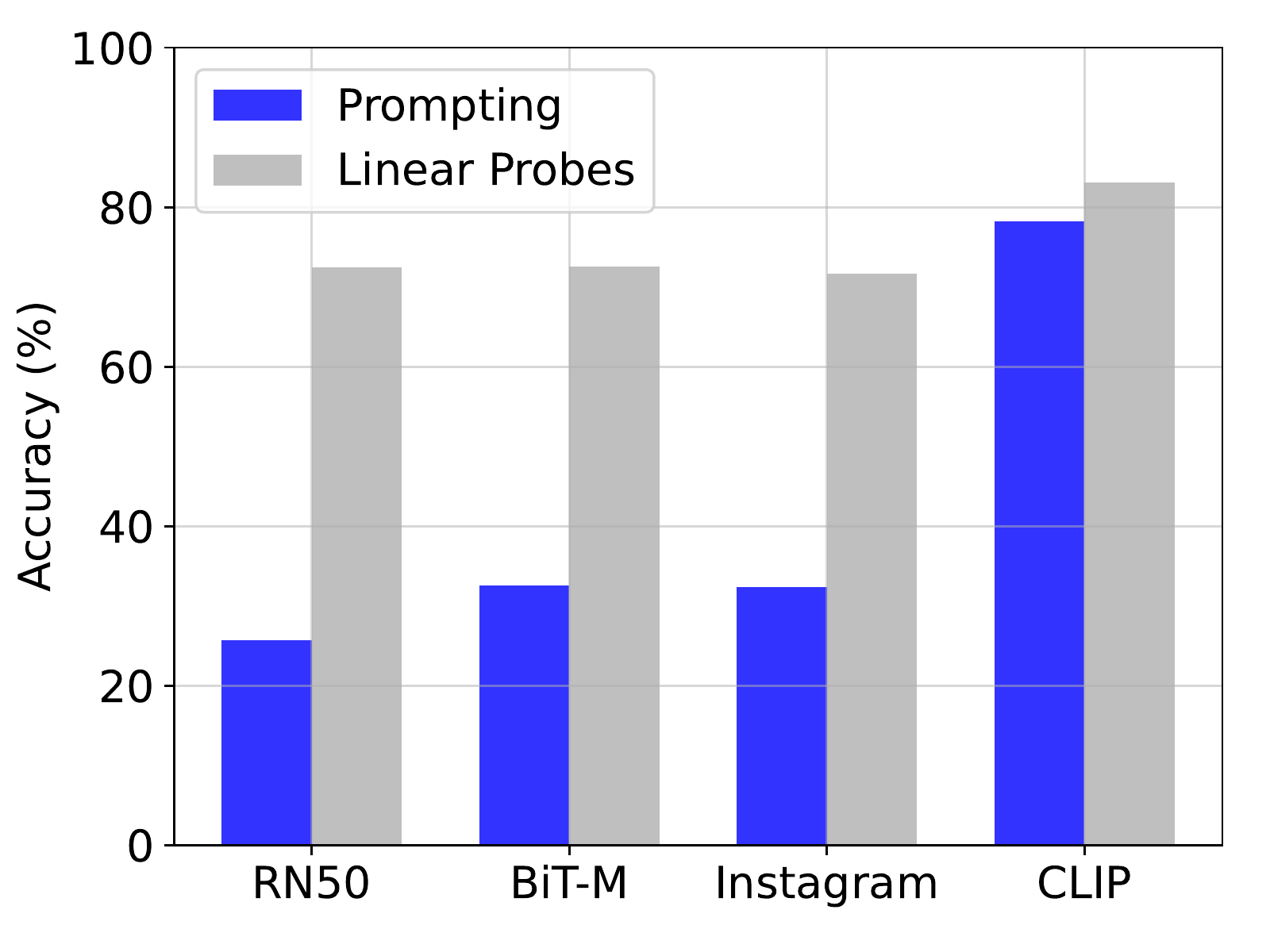}
\caption{\textbf{Prompting with vision models vs. CLIP}. Prompting with vision models shows significant performance gap to linear probe. In contrast, prompting with CLIP achieves competitive performance.}  
\label{result_summary}
    \end{minipage}\qquad
    \begin{minipage}[t]{.49\linewidth}
    \includegraphics[width=1\linewidth]{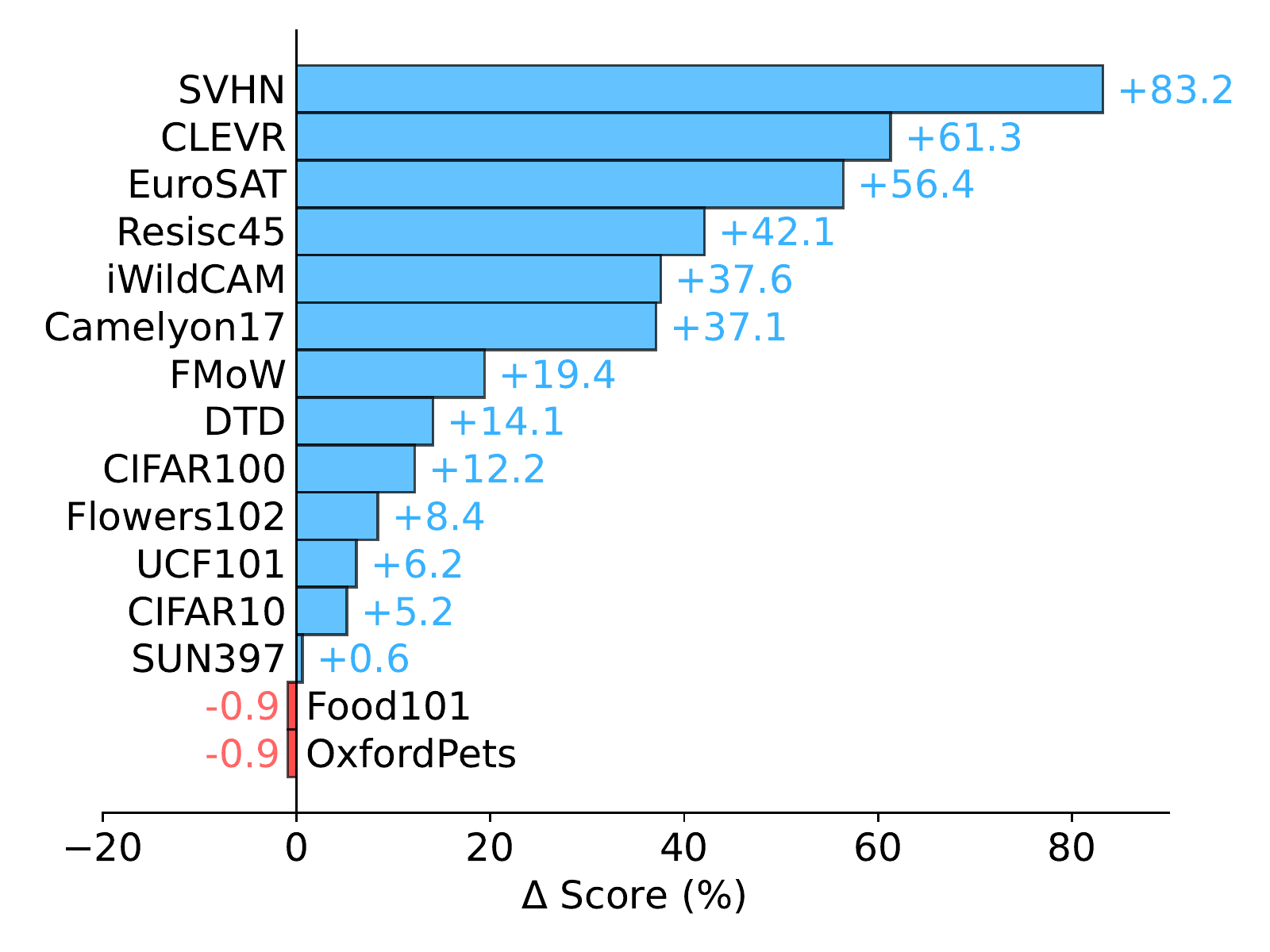}
\caption{\textbf{Accuracy gain from learning a visual prompt}. The bars indicate the gain (or loss) in accuracy obtained by learning a single visual prompt compared to text-prompted (i.e., zero-shot) CLIP.}
\label{delta_clip}
    \end{minipage}
\end{figure}

\begin{table}[t]
\caption{Performance across 12 datasets using CLIP. TP, VP, LP, and FT refer to text prompt, visual prompt, linear probe, and fine-tuning respectively. The green shade indicates cases where visual prompting outperforms linear probe.} 
\vskip -0.1in
\label{iid_clip}
\setlength\tabcolsep{0.2em}
\begin{center}
\begin{small}
\begin{tabular}{lccccccccccccccc}
\toprule
Model & Method & \scriptsize CIFAR100 & \scriptsize CIFAR10 & \scriptsize Flowers & \scriptsize Food & \scriptsize EuroSAT & \scriptsize SUN & \scriptsize UCF & \scriptsize SVHN & \scriptsize Pets & \scriptsize DTD & \scriptsize RESISC & \scriptsize CLEVR & Average \\
\midrule
CLIP & TP & 63.1 & 89.0 & 61.9 & 79.8 & 40.0 & 60.0 & 59.9 & 5.1 & 85.9 & 43.0 & 42.4 & 20.2 & 54.2\\
% \midrule
CLIP & VP + TP & 75.3 & 94.2 & 70.3 & 78.9 & \cellcolor{green!25}96.4 & 60.6 & 66.1 & \cellcolor{green!25}88.4 & 85.0 & 57.1 & 84.5 & \cellcolor{green!25}81.4 & 78.2\\
\midrule 
CLIP & LP & 80.0 & 95.0 & 96.9 & 84.6 & 95.3 & 75.0 & 83.3 & 65.4 & 89.2 & 74.6 & 92.3 & 66.0 & 83.1\\
\midrule 
CLIP & FT & 82.1 & 95.8 & 97.4 & 80.5 & 97.9 & 64.0 & 80.9 & 95.7 & 88.5 & 72.3 & 93.3 & 94.4 & 86.9\\
\bottomrule
\end{tabular}
\end{small}
\end{center}
\vskip -0.1in
\end{table}

\section{Results}

\subsection{Effectiveness of CLIP} 
We first compare prompting performance with linear probe, the current de facto approach to lightweight adaptation. Figure~\ref{result_summary} shows average test accuracy across 12 datasets for each pre-trained model. Prompting with vision models, or adversarial reprogramming, shows significant performance gap (+40\%) to standard linear probe. On the other hand, we find that prompting is surprisingly effective for CLIP, achieving competitive performance to linear probe. In particular, prompting outperforms linear probe on EuroSAT, SVHN, and CLEVR, by 1.1\%, 23\%, and 15.4\% respectively (Table~\ref{iid_clip}). On average, learning a visual prompt achieves 24\% performance gain compared to using text prompt only (i.e., ``zero-shot transfer''). Interestingly, we find that the performance of visual prompts varies across datasets (Figure~\ref{delta_clip}). Regarding this phenomenon, we further analyze what properties of the dataset affect performance in Section~\ref{dataset_property}. We report full results across 12 datasets for vision models in the Appendix.

\subsection{Robustness to Distribution Shift} 
As model parameters remain frozen, prompting prevents modifying the general knowledge base of the pre-trained model. This reduces the possibility of overfitting to spurious correlations in the downstream dataset, thereby improving robustness to distribution shift. Using the WILDS benchmark~\cite{wilds2021}, we learn visual prompts from training sets that contain images from a particular domain, and see how it transfers to test sets from different domains (e.g., images from different hospitals, regions, years, cameras). Table~\ref{ood_clip} show that average performance gap compared to linear probe and fine-tuning is further reduced to 4.5\% and 3.5\% respectively. On Camelyon17, visual prompting outperforms both linear probe and fine-tuning by 4.9\% and 6.5\% respectively. This suggests the practical utility of prompting in real-world deployments, where diverse range of domain shifts naturally arise. We report robustness results for vision models in the Appendix.

\begin{table}[h]
\begin{center}
\caption{Out-of-distribution test accuracy using CLIP.}
\label{ood_clip}
\setlength\tabcolsep{0.3em}
\begin{small}
\begin{tabular}{lccccc}
\toprule
Model & Method & \scriptsize iWILDCAM & \scriptsize FMoW & \scriptsize Camelyon17 & Average \\
\midrule
CLIP & TP & 14.1 & 13.5 & 52.7 & 26.8\\
% \midrule
CLIP & VP + TP & 51.7 & 32.9 & \cellcolor{green!25}89.8 & 58.1\\
\midrule
CLIP & LP & 66.7 & 36.3 & 84.9 & 62.6\\
\midrule
CLIP & FT & 54.9 & 46.6 & 83.3 & 61.6\\
\bottomrule
\end{tabular}
\end{small}
\end{center}
\vskip -0.1in
\end{table}

\section{Understanding Visual Prompts}
In this section, we investigate visual prompting performance in regard to properties of the downstream dataset, prompt design (i.e., input transformation), and output transformation.

\subsection{Downstream Dataset}
\label{dataset_property}
\begin{figure}[t]
    \begin{minipage}[t]{.49\linewidth}
    \centering
    \includegraphics[width=1\linewidth]{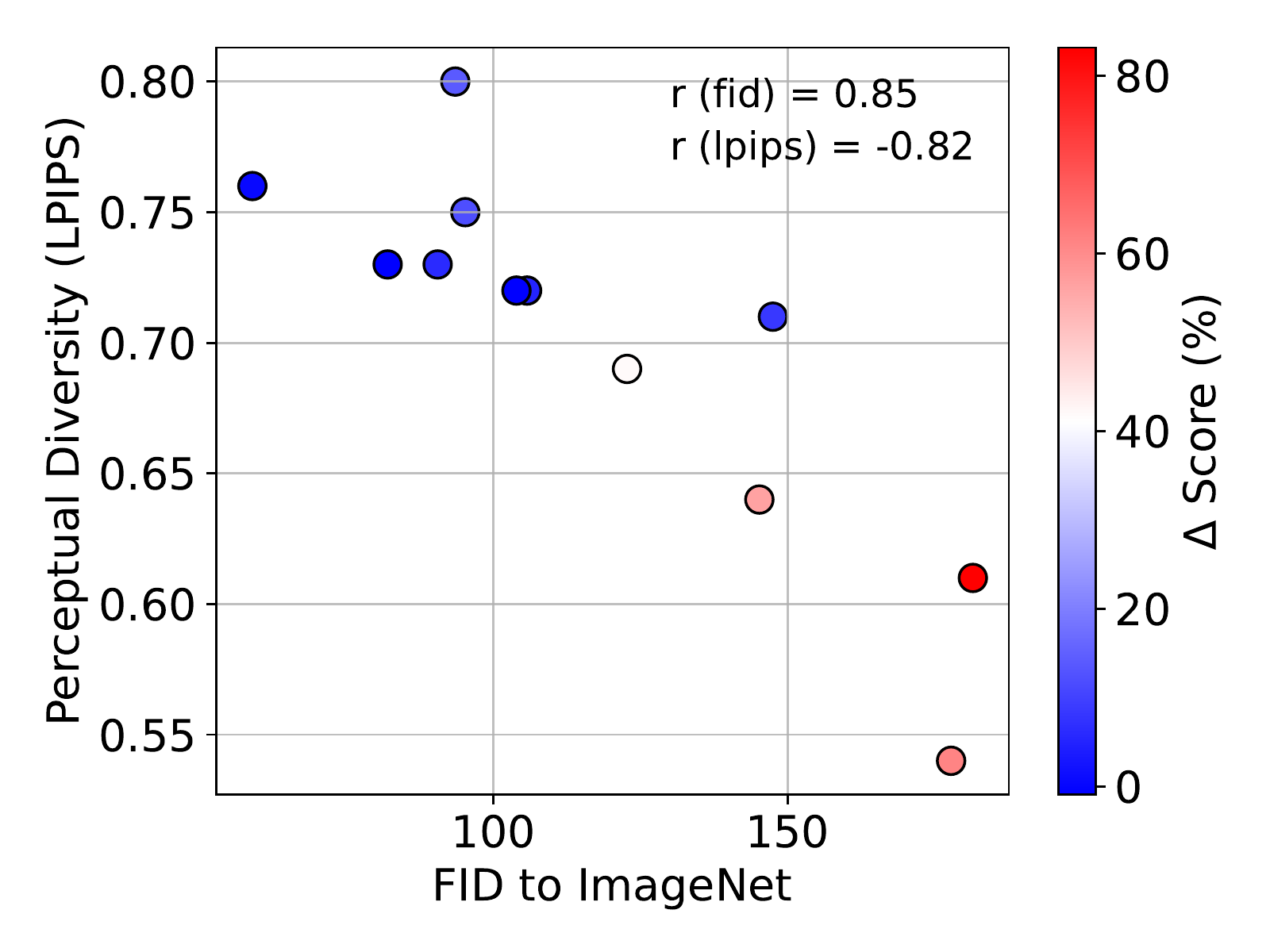}
\caption{\textbf{What properties of the downstream dataset affect performance?} We see a general performance gain as (1) datasets become more out-of-distribution to ImageNet, and (2) perceptual diversity of a dataset decreases.}.                                                           
\label{FID}
    \end{minipage}\qquad
    \begin{minipage}[t]{.49\linewidth}
    \includegraphics[width=1\linewidth]{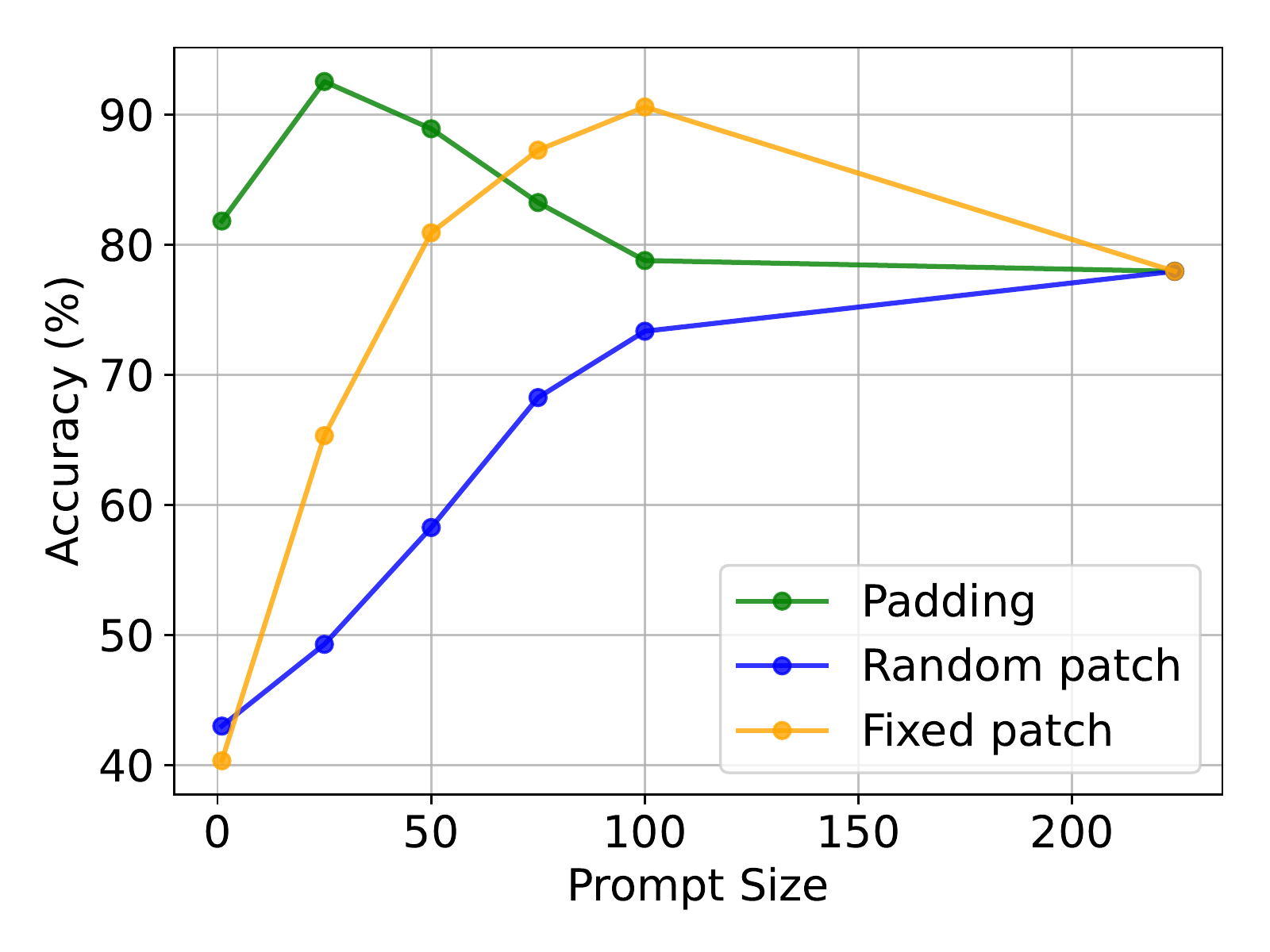}
\caption{\textbf{How does prompt design affect performance?} Using a moderate-size, padding template achieves best results on image classification tasks.}
\label{prompt_ablation}
\end{minipage}
\end{figure}

We find that the performance of visual prompting varies across downstream datasets. As shown in Figure~\ref{delta_clip}, the best-performing dataset achieves +83.2\% accuracy gain, while the worst-performing dataset has -1\% accuracy loss. To explain this phenomenon, we first hypothesize that visual prompts bridge the distribution gap by converting the \textit{unfamiliar} downstream dataset to look more similar to the pre-trained dataset. 
Under this hypothesis, visual prompts would not help datasets already within the pre-trained distribution, yet could help datasets that are severely out-of-distribution. While CLIP's pre-trained dataset is not available to the public, it is excessively tuned to achieve state-of-the-art zero-shot performance on ImageNet. Thus, we use ImageNet as a proxy. We validate our hypothesis by measuring the distributional similarity between ImageNet and downstream datasets using the FID score~\cite{heusel2017gans}. We compare these scores to accuracy gain from visual prompts. Due to computation limitations, we randomly sample 100k images from the ImageNet-1k training set to compute the metrics. In Figure~\ref{FID}, we observe a general performance gain as the downstream dataset becomes more out-of-distribution to ImageNet (e.g., CLEVR, SVHN).

Another hypothesis regards to learning a single prompt per dataset. While this may be sufficient for datasets with low perceptual diversity, a single visual prompt may fail to capture the full distribution as the diversity increases. We measure perceptual diversity using LPIPS~\cite{zhang2018unreasonable}. For each dataset, we measure LPIPS between two randomly sampled image pairs and report the average score. Figure~\ref{FID} shows that a learning single visual prompt achieves better performance gain for datasets with low perceptual diversity. 

\subsection{Prompt Design}
\label{prompt_ablation}

\begin{wrapfigure}{r}{0.3\textwidth}
\begin{center}
\vspace{-0.2in}
\includegraphics[width=0.23\textwidth]{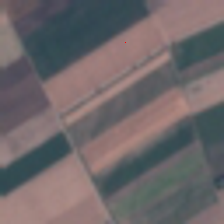}
  \end{center}
  \vspace{-0.1in}
  \caption{Given a frozen CLIP, adding a \emph{single} pixel to EuroSAT achieves +3\% accuracy (zoom in to find the red pixel).}
  \label{single_pixel}
  \vspace{-0.1in}
\end{wrapfigure}

Choosing the right prompt design (i.e., template and size) can highly affect performance. We perform an ablation study on three different templates: pixel patch at random location, pixel patch at fixed location, and padding, across prompt size $p=1, \dots, 224$. We measure accuracy on the EuroSAT dataset using a frozen CLIP. Figure~\ref{prompt_ablation} shows that using a fixed-location template (i.e., padding, fixed patch) yields better performance. For fixed-location templates, we find that performance improves as prompt size increases (i.e., more trainable parameters), then it starts to drop for +70k parameters. Surprisingly, we find that our simplest approach --- adding a \textit{single}-pixel prompt --- can yield a 3\% improvement over text-prompted CLIP (Figure~\ref{single_pixel}). Overall, padding with $p=30$ achieves the best performance in our experiments. We believe this is because our application scope is image classification, where the object of interest tends to be located in the center of the image. We believe other visual tasks may require significantly different design choices. Refer to Section~\ref{prompt_design} on how the actual number of parameters are calculated.

\subsection{Output Transformation}
\label{text_quality}
We investigate prompting performance in regard to how we design the output transformation. For vision models, we follow~\cite{elsayed2018adversarial} and use hard-coded mapping; downstream class indices are arbitrarily assigned to pre-trained class indices. We analyze how this mapping affects downstream performance. Using a subset of OxfordPets, we construct a simple toy dataset for classifying dogs and cats. Using ResNet trained on ImageNet-1k (RN50), we compare two cases: (1) downstream classes are assigned to pre-trained classes with similar semantics (unseen ``dog'' assigned to pre-trained ``chihuahua'' index), (2) we swap the indices (cat assigned to dog index, vice versa). (1) achieves 100\% and (2) achieves 62.5\%; having similar semantics between class indices is critical for performance. This may explain the performance gap between vision models and CLIP. 

\begin{wrapfigure}{r}{0.5\textwidth}
\begin{center}
\vspace{-0.2in}
\includegraphics[width=0.49\textwidth]{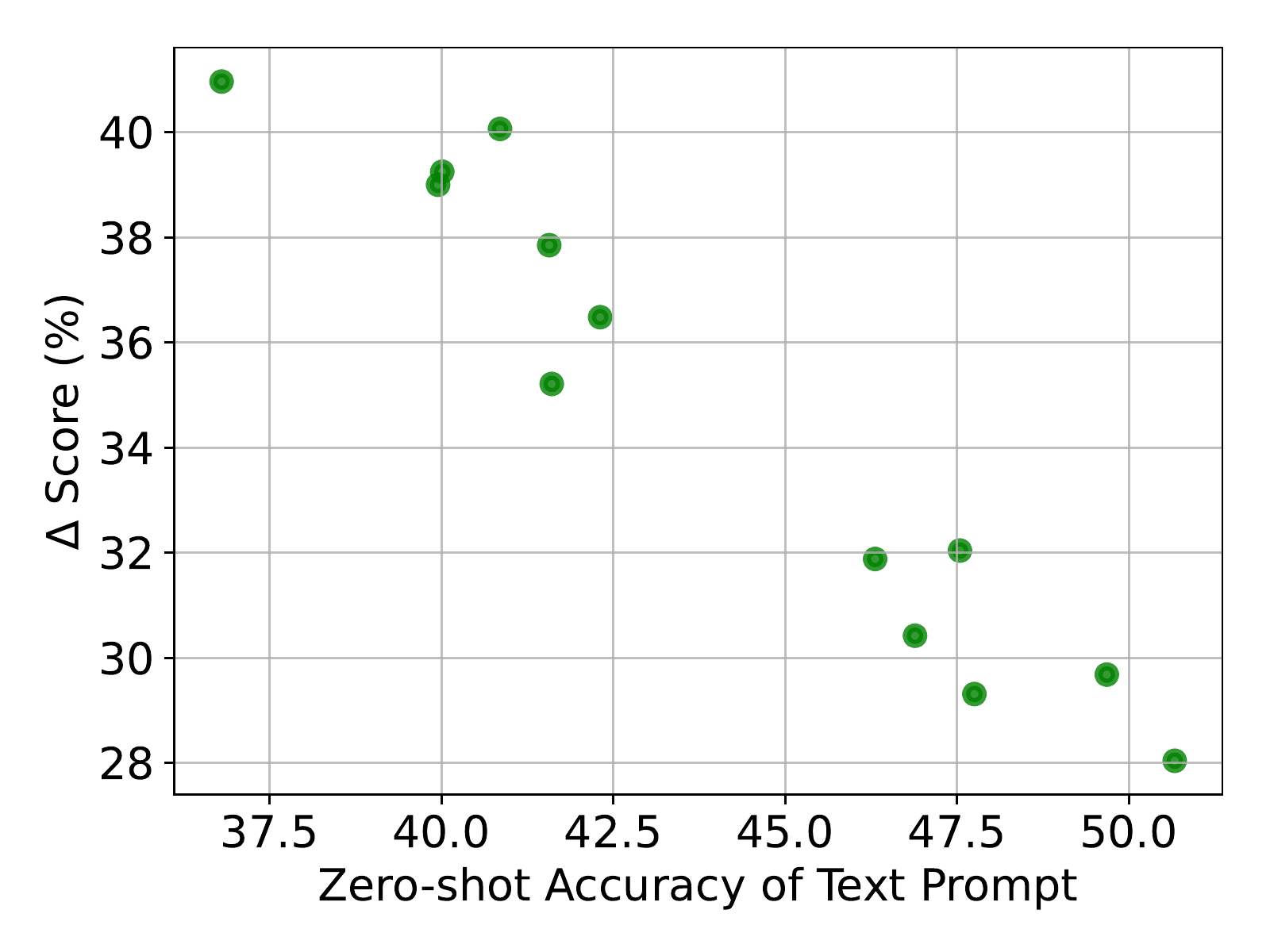}
  \end{center}
  \vspace{-0.1in}
\caption{\textbf{Visual prompts compensate for low-quality text prompts}. We measure the quality of text prompts with the zero-shot accuracy of CLIP. As zero-shot accuracy decreases (i.e., lower quality text prompt), accuracy gain from visual prompting increases.} 
\label{text_qual}
  \vspace{-0.1in}
\end{wrapfigure}

For CLIP, a vision-language model, we use text prompts for output transformation. As we learn visual prompts via backpropagation, the learning signal is dependent on the text prompt we use. It has been reported that CLIP's zero-shot accuracy can be significantly improved by using a better text prompt~\cite{radford2021learning}. Therefore, we hypothesize that the quality of text prompt affects the performance of visual prompts. On EuroSAT, we measure text prompt quality by the zero-shot performance of CLIP. Figure~\ref{text_qual} shows that the performance gain from visual prompts is higher for text prompts with low zero-shot performance. In other words, visual prompting can compensate for low-quality text prompts. As manually searching for the best text prompt is extremely laborsome, this result highlights the usefulness of visual prompts.

\section{Discussion}
In this paper, we have investigated a method to perturb inputs to a pre-trained model in a manner which
improves classification accuracy. A broader interpretation of visual prompting is to think of it as a way to steer a pre-trained model in \textit{any} direction by modifying its input space. For instance, a visual prompt for an image-to-image model could be used to change the visual style of the input. Even though we have explored ``universal'' visual prompts in this work (i.e., a single prompt that apply to all input images), prompts could also be made input-conditional and hence less universal but perhaps more accurate. The specific design choices including (a) input-specific or input-agnostic, (b) improving or decreasing accuracy, and (c) type of the pretrained-model, can be modified to create future interesting applications of prompting. 

% \begin{wrapfigure}{r}{0.3\textwidth}
% \includegraphics[width=0.2\textwidth]{figures/text_qual.pdf}
% \caption{\textbf{Visual prompts compensate for low-quality text prompts}. We measure the quality of text prompts with the zero-shot accuracy of CLIP. As zero-shot accuracy decreases (i.e., lower quality text prompt), accuracy gain from visual prompting increases.} 
% \label{text_qual}
% \end{wrapfigure}

One natural question that arises following our exposition is in what situations would one prefer visual prompting over fine-tuning or a linear probe? Fine-tuning assumes that the model can be modified which may not always be the case (e.g., if the model is exposed by a API owned by a third-party). While prompting does under-perform linear probe in some cases, we would like to stress that the goal of this work is to show the existence of a prompting mechanism in ``pixel space'', which works across multiple datasets and pre-trained models, and reveals new avenues for how vision models can be effectively adapted. Our focus in this work is not to outperform state-of-the-art; we note that there are several approaches one could use to improve performance further including ensembling multiple prompts, using prompts in conjunction with linear probe or fine-tuning, or scaling the pre-trained model (e.g., ViT-L/14 of CLIP, which is unfortunately not available to the public). We leave these for future work.

\section{Conclusion}
While standard adaptation methods in vision focus on introducing a separate task-specific head and adapt the model parameters or activations, we investigate visual prompting as practical adaptation method. We use a gradient-based scheme to learn a single, input-agnostic perturbation that repurposes a frozen model to perform a downstream task. Through various experiments across pre-trained models and datasets, we have demonstrated that CLIP is particularly suitable for visual prompting, achieving competitive results to linear probe. We hope that our unique findings will spur further research into: (1) better understanding pixel-space adaptation --- when and why they are effective at steering deep networks, and (2) developing better visual prompts that further add to our repertoire of mechanisms for creating flexible and adaptable vision systems.

\section*{Acknowledgements}
We would like to thank Lucy Chai, Caroline Chan, Joanna Materzynska, Xavier Puig Fernandez, Minyoung Huh, Tongzhou Wang, and Yen-Chen Lin for proofreading the paper. We thank Judy Hoffman for helpful discussion and advice. This work was partially supported by funding from MIT STL and an MIT RSC award from the NEC fund. 

% %%%%%%%%%%%%%%%%%%%%%%%%%%%%%%%%%%%%%%%%%%%%%%%%%%%%%%%%%%%%

\bibliographystyle{plain}
\bibliography{neurips_2022}

%%%%%%%%%%%%%%%%%%%%%%%%%%%%%%%%%%%%%%%%%%%%%%%%%%%%%%%%%%%%

\newpage
\appendix

\section{Appendix}

\begin{table*}[h]
\caption{Overview of pre-trained models.}
\vskip -0.1in
%\vspace*{-\baselineskip}
\label{pretrain_models}
\vskip 0.15in
\begin{center}
\begin{small}
\begin{tabular}{ccccc}
\toprule
Model & Architecture & Modality & Pre-trained Dataset & Objective\\
\midrule
CLIP~\cite{radford2021learning} & ViT-B/32 & Vision-language & 400M image-text pairs & Contrastive \\
Instagram~\cite{mahajan2018exploring} & ResNext101-32x8d & Vision & 3.5B Instagram photos & Cross Entropy \\
BiT-M~\cite{kolesnikov2020big} & ResNet-50 & Vision & 14M ImageNet-21k & Cross Entropy\\
RN50~\cite{he2016deep} & ResNet-50 & Vision & 1.2M ImageNet-1k & Cross Entropy\\
\bottomrule
\end{tabular}
\end{small}
\end{center}
\end{table*} 

\begin{table*}[!htb]
\caption{Performance across 12 datasets using vision pre-trained models. TP, VP, LP, and FT refer to text prompt, visual prompt, linear probe, and fine-tuning respectively. The green shade indicates cases where visual prompting outperforms linear probe.}
\vspace*{-\baselineskip}
\label{iid_vision}
\vskip 0.15in
\setlength\tabcolsep{0.3em}
\begin{center}
\begin{small}
\begin{tabular}{lccccccccccccccc}
\toprule
Model & Method & \scriptsize CIFAR100 & \scriptsize CIFAR10 & \scriptsize Flowers & \scriptsize Food & \scriptsize EuroSAT & \scriptsize SUN &\scriptsize UCF & \scriptsize SVHN & \scriptsize Pets &\scriptsize DTD & \scriptsize RESISC & \scriptsize CLEVR & Average \\
\midrule
%Instagram & - & 1.5 & 7.0 & 1.5 & 0.8 & 7.2 & 0.3 & 0.5 & 12.7 & 1.2 & 2.1 & 2.4 & 15.4 & 4.4\\
Instagram & VP & 16.7 & 62.1 & 22.9 & 9.9 & 85.4 & 2.2 & 15.4 & \cellcolor{green!25}53.8 & 18.6 & 29.1 & 41.4 & \cellcolor{green!25}30.9 & 32.4\\
Instagram & LP & 64.0 & 90.1 & 92.7 & 65.8 & 90.6 & 58.1 & 76.6 & 48.0 & 94.5 & 70.9 & 79.2 & 30.2 & 71.7\\
Instagram & FT & 77.8 & 77.8 & 94.5 & 75.6 & 97.4 & 56.7 & 72.9 & 96.8 & 93.9 & 73.5 & 93.4 & 87.9 & 83.2\\
\hline & \\[-2.0ex]
%BiT-M & - & 0.7 & 9.6 & 0.4 & 0.7 & 16.0 & 0.2 & 0.9 & 6.1 & 4.2 & 2.6 & 3.2 & 13.3 & 4.8\\
BiT-M & VP & 16.2 & 53.6 & 29.2 & 11.6 & 72.5 & 2.6 & 16.6 & \cellcolor{green!25}56.3 & 24.7 & 30.0 & 43.0 & \cellcolor{green!25}34.8 & 32.6\\
BiT-M & LP & 73.0 & 90.8 & 99.1 & 72.6 & 94.4 & 49.5 & 72.9 & 49.8 & 85.2 & 68.4 & 87.7 & 28.4 & 72.6\\
BiT-M & FT & 76.2 & 94.1 & 99.4 & 75.6 & 98.3 & 52.7 & 81.3 & 97.2 & 89.0 & 69.3 & 93.6 & 88.2 & 84.6 \\
\hline & \\[-2.0ex]
%RN50 & - & 1.5 & 10.0 & 1.6 & 0.8 & 7.7 & 0.3 & 1.3 & 21.6 & 0.9 & 2.2 & 4.6 & 11.7 & 5.3\\
RN50 & VP & 10.1 & 54.5 & 14.0 & 5.1 & 78.7 & 1.1 & 9.5 & 57.1 & 10.8 & 8.2 & 29.9 & 29.5 & 25.7\\
RN50 & LP & 67.7 & 87.7 & 92.7 &62.5 & 94.5 &57.5 &69.4 & 60.3 & 91.1 &66.7 & 87.1 & 32.6 & 72.5\\
RN50 & FT & 79.9 & 94.1 & 96.9 & 73.2 & 96.5 & 55.9 & 76.7 & 96.9 & 92.3& 66.7 & 93.4 & 89.3 & 84.3\\
\bottomrule
\end{tabular}
\end{small}
\end{center}
\end{table*}

\begin{table}[h]
\caption{Out-of-distribution test accuracy on WILDS using vision pre-trained models. The green shade indicates a case where visual prompting outperforms linear probe and fine-tuning.}
\vspace*{-\baselineskip}
\vskip 0.05in
\label{ood_vision}
\setlength\tabcolsep{0.3em}
\begin{center}
\begin{small}
\begin{tabular}{lccccc}
\toprule
Model & Method & \scriptsize iWILDCAM & \scriptsize FMoW & \scriptsize Camelyon17 & Average \\
\midrule
% \midrule
%Instagram & - & 0.1 & 1.6 & 49.9 & 17.2\\
Instagram & VP & 52.2 & 14.0 & \cellcolor{green!25}87.1 & 51.1\\
Instagram & LP & 64.1 & 22.7 & 77.4 & 54.7\\
Instagram & FT & 62.6 & 46.8 & 86.5 & 65.3\\
\hline & \\[-2.0ex]
%BiT-M & - & 0.0 & 1.6 & 50.0 & 17.2\\
BiT-M & VP & 48.7 & 15.7 & 83.3 & 49.2\\
BiT-M & LP & 55.9 & 22.3 & 89.0 & 55.7\\
BiT-M & FT & 59.0 & 49.8 & 81.8 & 63.5 \\
\hline & \\[-2.0ex]
%RN50 & - & 0.3 & 1.6 & 55.4 & 19.1\\
RN50 & VP & 51.9 & 12.6 & 84.5 & 49.7\\
RN50 & LP & 62.7 & 28.7 & 90.2 & 60.5\\
RN50 & FT & 62.2 & 46.7 & 88.0 & 65.6\\
\bottomrule
\end{tabular}
\end{small}
\end{center}
% \vskip -0.1in
\end{table}

\begin{figure}[!htb]
%\vskip 0.2in
\begin{center}
\centerline{\includegraphics[width=1\textwidth]{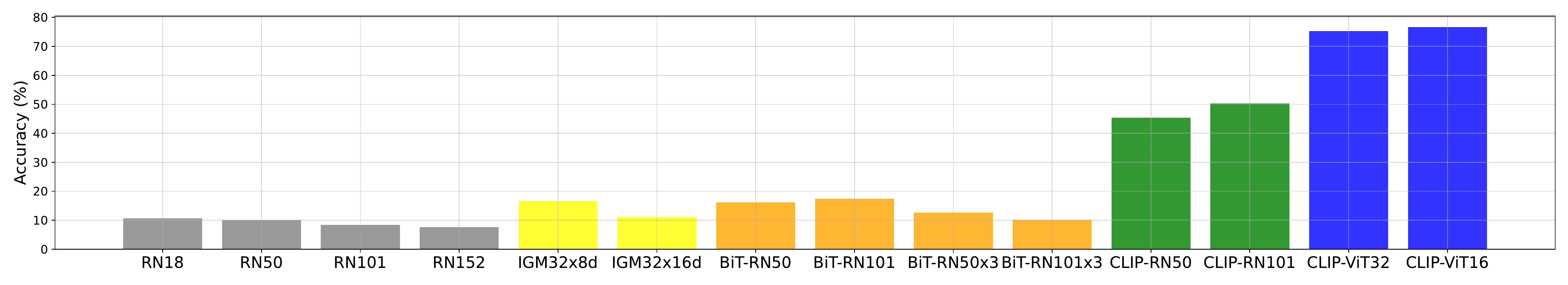}}
% \vskip -0.1in
\caption{Model architecture ablation on CIFAR100.}
\label{model_ablation}
\end{center}
\vskip -0.2in
\end{figure}

\subsection{Ablation on Model Architecture}
In Figure~\ref{model_ablation}, we compare performance using different model architectures on CIFAR100. We compare the original ImageNet-pretrained ResNets released by~\cite{he2016deep}, namely ResNet-18, ResNet-50, ResNet-101, ResNet-152. For Instagram-pre-trained ResNeXt~\cite{mahajan2018exploring}, we compare two models (32x8d, 32x16d). For Big Transfer~\cite{kolesnikov2020big}, we use four BiT-M models (ResNet-50, ResNet-101, ResNet-50x3, ResNet-101x3). For ResNet-based CLIP models, we compare two models trained on $224\times224$ images (ResNet-50, ResNet-101). For CLIP models that use the Vision Transformer~\cite{dosovitskiy2020image}, we compare the two released models (ViT-B/32, ViT-B/16).
For vision models, performance does not necessarily increase for larger models. For CLIP, we observe superiority of ViT-based models over ResNet-based models.

\begin{figure*}[!htb]
\vskip 0.2in
\begin{center}
\centerline{\includegraphics[width=1\textwidth]{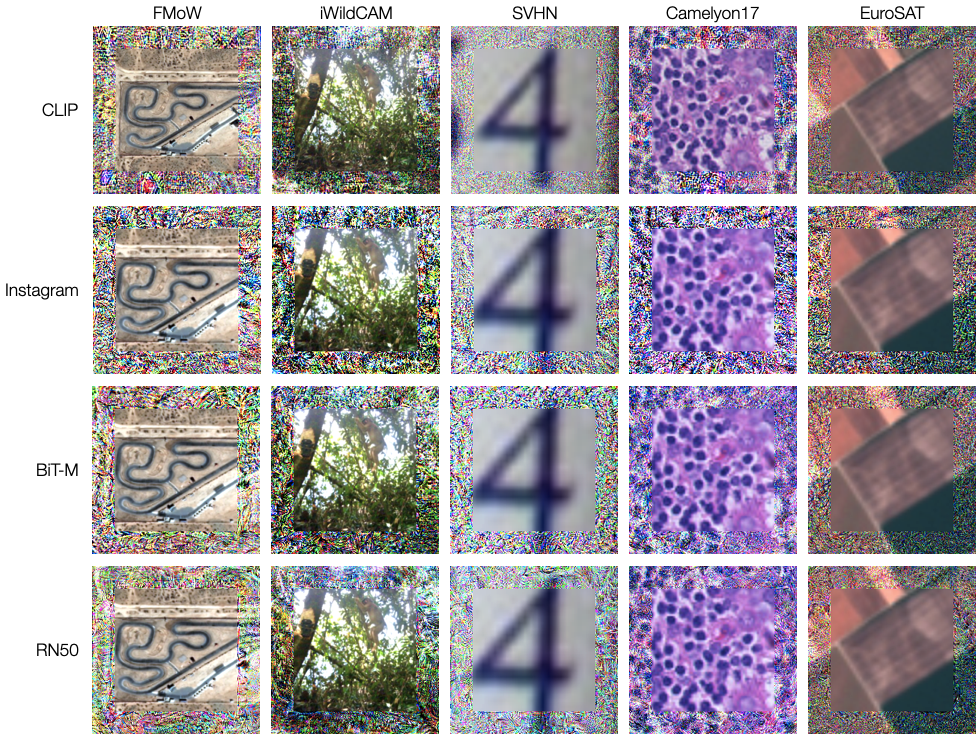}}
\caption{Visualizing task-specific visual prompts for each pre-trained model.}
\label{visualize}
\end{center}
\vskip -0.2in
\end{figure*}
% In Figure~\ref{visualize}, we visualize the prompts for different datasets and pre-trained models. While uninterpretable to human eye, adding this pixel patch to every test image allows the frozen model to adapt better to a new task. We believe visual prompting 
% aligns pre-trained and downstream distributions by converting the downstream dataset to look more similar to the pre-trained dataset. Under this hypothesis, visual prompting would not help for datasets already within the pre-trained distribution, yet could help significantly for datasets that are severely out-of-distribution, which may explain the performance gap between tasks.  

% \subsection{Dataset Statistics}
% Table~\cite{}
\begin{table}[!htb]
\caption{Description of the datasets and the corresponding text prompt used for CLIP.}
\vspace*{-\baselineskip}
\setlength\tabcolsep{0.3em}
\begin{center}
\begin{small}
\begin{tabular}{cccccc}
\toprule
Dataset & Train Size & Validation Size & Test Size & Classes & Text Prompt \\
\midrule
CIFAR100 & 50,000 & - & 10,000 & 100 & ``This is a photo of a \{ \}''\\
CIFAR10 & 50,000 & - & 10,000 & 10 & ``This is a photo of a \{ \}''\\
Flowers102 & 4,093 & 1,633 & 2,463 & 102 & ``This is a photo of a \{ \}''\\
Food101 & 50,500 & 20,200 & 30,300 & 101 & ``This is a photo of a \{ \}''\\
EuroSAT & 13,500 & 5,400 & 8,100 & 10 & ``This is a photo of a \{ \}''\\
SUN397 & 15,888 & 3,970 & 19,850 & 397 & ``This is a photo of a \{ \}''\\
UCF101 & 7,639 & 1,898 & 3,783 & 101 & ``This is a photo of a \{ \}''\\
SVHN & 73,257 & - & 26,032 & 10 & ``This is a photo of a \{ \}''\\
OxfordPets & 2,944 & 736 & 3,669 & 37 & ``This is a photo of a \{ \}''\\
DTD & 2,820 & 1,128 & 1,692 & 47 & ``This is a photo of a \{ \}''\\
Resisc45 & 18,900 & 6,300 & 6,300 & 45 & ``This is a photo of a \{ \}''\\
CLEVR/count & 70,000 & - & 15,000 & 8 & ``This is a photo of \{ \} objects'' \\
iWildCAM & 129,809 & 14,961 & 42,791 & 182 & ``This is a photo of a \{ \}''\\
FMoW & 76,863 & 19,915 & 22,108 & 62 & ``This is a photo of a \{ \}''\\
Camelyon17 & 302,436 & 34,904 & 85,054 & 2 & ``a tissue region \{ \} tumor'' \\
\bottomrule
\end{tabular}
\end{small}
\end{center}
\label{data_stats}
\vskip -0.1in
\end{table}

\subsection{Dataset Statistics}
Table~\ref{data_stats} illustrates description of the datasets and the corresponding text prompt used for adapting CLIP. For OxfordPets, Flowers102, Food101, SUN397, DTD, EuroSAT, and UCF101, we used the data splits provided by~\cite{zhou2021learning}. For other datasets, we used the officially provided data splits.

\subsection{Change Log}
\paragraph{ArXiv v2} In ArXiv v1, we overlooked the adversarial reprogramming literature. In fact, visual prompting for vision models is essentially the same as adversarial reprogramming! In the current version, we have clarified this and removed claims of methodological novelty. We have reframed the paper as an exploration of the viability of visual prompts as a practical adaptation method for modern large-scale models. We thank Seong Joon Oh and users on twitter for pointing out the connection to adversarial reprogramming.

\end{document}